\documentclass[Afour,sageh,times]{sagej}

\usepackage{moreverb,url}

\usepackage[colorlinks,bookmarksopen,bookmarksnumbered,citecolor=red,urlcolor=red]{hyperref}
\usepackage{orcidlink} 

\usepackage{colortbl} 
\usepackage{booktabs} 
\usepackage{siunitx}
\usepackage{etoolbox}
\usepackage{makecell} 
\usepackage{multirow}

\newcommand\BibTeX{{\rmfamily B\kern-.05em \textsc{i\kern-.025em b}\kern-.08em
T\kern-.1667em\lower.7ex\hbox{E}\kern-.125emX}}

\def\volumeyear{2024}

\begin{document}

\runninghead{De Gusseme et al.}

\title{A Dataset and Benchmark for Robotic Cloth Unfolding Grasp Selection: The ICRA 2024 Cloth Competition}

\author{
Victor-Louis De Gusseme\orcidlink{0000-0003-2046-573X}, %
Thomas Lips\orcidlink{0000-0001-9530-5349},
Remko Proesmans\orcidlink{0000-0002-5925-625X},
Julius Hietala\orcidlink{0009-0006-4147-2313}, 
Giwan Lee,
Jiyoung Choi,
Jeongil Choi,
Geon Kim,
Phayuth Yonrith,
Domen Tabernik\orcidlink{0000-0002-5613-5882},
Andrej Gams\orcidlink{0000-0002-9803-3593},
Peter Nimac\orcidlink{0000-0001-8690-0293},
Matej Urbas,
Jon Muhovič\orcidlink{0000-0002-4082-2506},
Danijel Skočaj\orcidlink{0000-0002-5290-4736},
Matija Mavsar\orcidlink{0000-0002-1617-4908},
Hyojeong Yu\orcidlink{0009-0007-8402-3494},
Minseo Kwon\orcidlink{0009-0001-9766-1870},
Young J. Kim\orcidlink{0000-0003-2159-4832},
Yang Cong\orcidlink{0000-0002-5102-0189}, 
Ronghan Chen\orcidlink{0000-0001-6307-2923}, 
Yu Ren\orcidlink{0000-0003-2637-9728}, 
Supeng Diao\orcidlink{0009-0003-3352-9445}, 
Jiawei Weng\orcidlink{0009-0007-7795-1639}, 
Jiayue Liu\orcidlink{0009-0007-1162-2971},
Haoran Sun\orcidlink{0009-0004-0090-2511}, 
Linhan Yang\orcidlink{0000-0001-7393-0741}, 
Zeqing Zhang\orcidlink{0000-0002-5935-3674}, 
Ning Guo\orcidlink{0000-0002-4963-3381}, 
Lei Yang\orcidlink{0000-0002-3284-4019}, 
Fang Wan\orcidlink{0000-0002-8658-1197}, 
Chaoyang Song\orcidlink{0000-0002-0166-8112}, 
Jia Pan\orcidlink{0000-0001-9003-2054},
Yixiang Jin\orcidlink{0000-0001-6286-278X}, 
Yong A\orcidlink{354018 0009-0005-8709-9027}, 
Jun Shi\orcidlink{353919 0009-0007-3770-6663}, 
Dingzhe Li\orcidlink{291827 0009-0007-7520-0620}, 
Yong Yang,
Kakeru Yamasaki\orcidlink{0009-0003-6777-6903},
Takumi Kajiwara,
Yuki Nakadera,
Krati Saxena\orcidlink{0000-0001-7049-9685},
Tomohiro Shibata\orcidlink{0000-0002-8766-4250},
Chongkun Xia\orcidlink{0000-0001-5396-7643},
Kai Mo\orcidlink{0000-0003-1269-978X},
Yanzhao Yu\orcidlink{0009-0009-5958-5887},
Qihao Lin\orcidlink{0009-0001-3055-6208},
Binqiang Ma\orcidlink{0009-0003-8136-7823},
Uihun Sagong\orcidlink{0000-0002-1602-378X}, 
JungHyun Choi\orcidlink{0000-0001-5814-2392},
JeongHyun Park\orcidlink{0000-0001-8139-6024}, 
Dongwoo Lee\orcidlink{0009-0006-3844-8443}, 
Yeongmin Kim\orcidlink{0009-0000-1178-8612}, 
Myun Joong Hwang\orcidlink{0000-0003-1272-9985}
Yusuke Kuribayashi\orcidlink{0009-0009-0470-5773},
Naoki Hiratsuka,
Daisuke Tanaka\orcidlink{0000-0002-3489-5792},
Solvi Arnold\orcidlink{0000-0003-4342-9344},
Kimitoshi Yamazaki\orcidlink{0000-0002-4096-3288},
Carlos Mateo-Agullo\orcidlink{0000-0002-1730-4233},
Andreas Verleysen\orcidlink{0000-0001-5071-5619},
Francis wyffels\orcidlink{0000-0002-5491-8349}
}



\corrauth{Francis wyffels, AI and Robotics Lab (IDLab-AIRO), Ghent University - imec, Ghent, Belgium.}

\email{francis.wyffels@ugent.be}

\begin{abstract}

Robotic cloth manipulation suffers from a lack of standardized benchmarks and shared datasets for evaluating and comparing different approaches.
To address this, we created a benchmark and organized the ICRA 2024 Cloth Competition, a unique head-to-head evaluation focused on grasp pose selection for in-air robotic cloth unfolding.  Eleven teams participated in the competition, utilizing the publicly released dataset of 500 real-world robotic grasp attempts for cloth unfolding and employing diverse approaches to generate in-air unfolding grasps. 
Analysis of the competition results revealed insights about the trade-off between grasp success and coverage, the surprisingly strong achievements of hand-engineered methods and a significant discrepancy between competition performance and prior work, underscoring the importance of independent, out-of-the-lab evaluation in robotic cloth manipulation. 
We also expanded the dataset with 176 competition evaluation trials, resulting in a dataset of 679 unfolding demonstrations across 34 garments.
This dataset is a valuable resource for developing and evaluating grasp selection methods, particularly for learning-based approaches.
 We hope that the benchmark, dataset and competition results can serve as a foundation for future benchmarks and drive further progress in data-driven robotic cloth manipulation.
\end{abstract}

\keywords{benchmarking, datasets, cloth manipulation, unfolding, robotics competition}

\maketitle

\section{Introduction}

\begin{figure*}[!t]
    \centering
    \includegraphics[width=\linewidth]{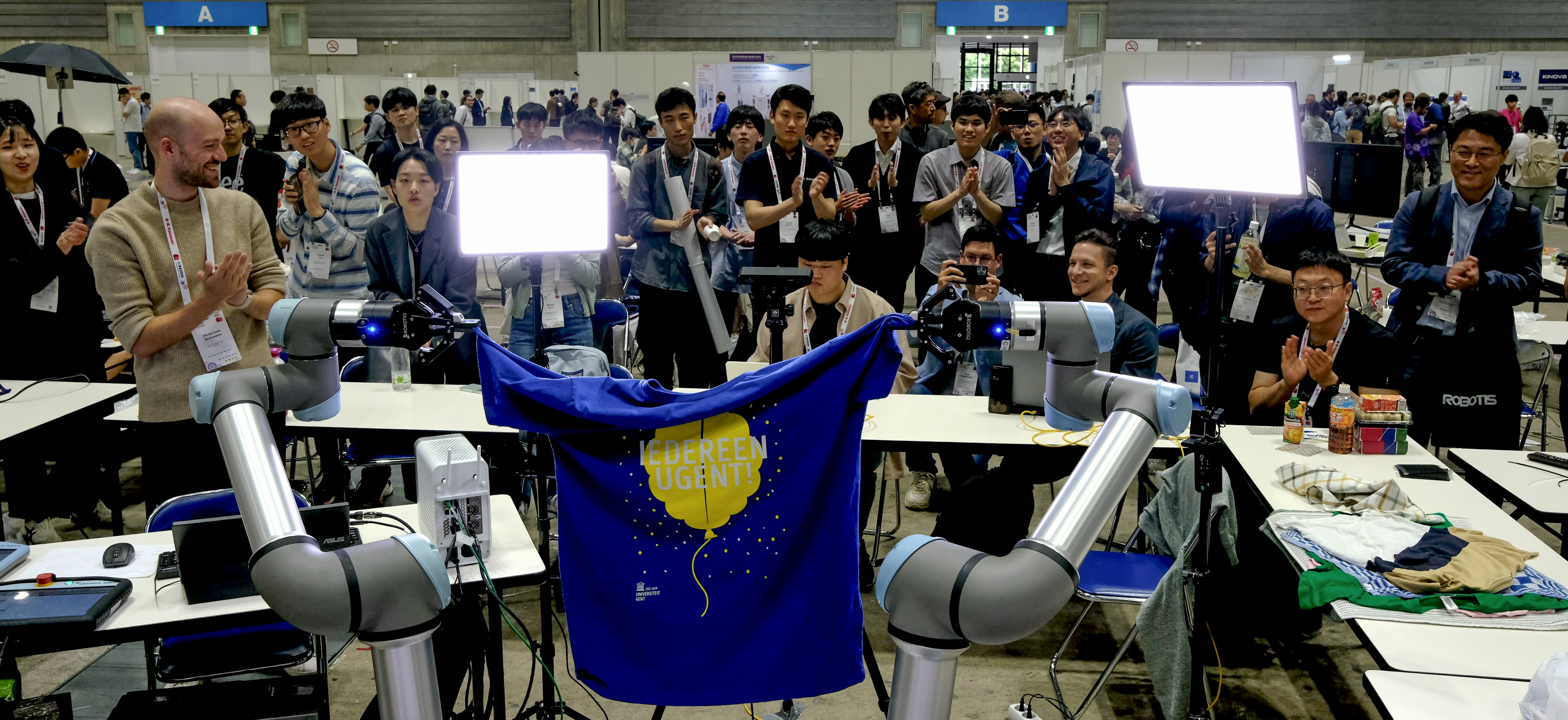}
    \caption{Impression of the ICRA 2024 Cloth Competition on in-air unfolding. The dual-arm robotic setup, surrounded by participants and spectators, executes the final stretching motion to achieve a nearly perfectly unfolded T-shirt. All competition evaluations were conducted on-site at the conference in Yokohama, Japan, using this standardized, shared setup.}
    \label{venue}
\end{figure*}

Datasets and benchmarks play a crucial role in the field of robotics, providing the means for objective evaluation, tracking progress through time, and ensuring reproducibility~\citep{calli2015benchmarking}.
They facilitate a shared understanding of the state-of-the-art, allow for analyzing remaining challenges and provide a means of interaction between individual laboratories.

Significant progress has been made in benchmarking rigid object manipulation~\citep{kimble2020benchmarking, Cruciani2020Benchmarking}, particularly in grasping and pick-and-place tasks~\citep{correll2018analysis, davella2023cluttered}.
However, the challenges inherent to deformable object manipulation~\citep{sanchez2018robotic, yin2021modeling, kadi2023datadriven}, especially with cloth~\citep{longhini2024unfolding}, have led to a comparative lack of shared datasets and standardized benchmarks~\citep{irene2020benchmarking}. The deformability of cloth poses unique challenges for reproducibility, necessitating dedicated efforts in this domain.

Building on previous work on benchmarking cloth manipulation~\citep{garcia2022clothcomp}, we present the IEEE International Conference on Robotics and Automation (ICRA) 2024 Cloth Competition and its associated dataset and benchmark.
The benchmark and competition focus on grasp selection algorithms for in-air robotic cloth unfolding, and the format was streamlined to focus the participant's efforts on grasp selection. 
Rigorous evaluation was prioritized by testing generalization to different cloth items, using objective and automated evaluation, and performing on-site, live evaluations on a shared robot platform.
The competition ultimately conducted 176 evaluations across 11 diverse teams.
This effort resulted in a comprehensive dataset of 679 real-world robotic cloth unfolding attempts across 34 diverse garments, making it one of the largest and most diverse datasets in this domain. The competition and the dataset enjoy a symbiotic relationship, with the competition illustrating the dataset's utility and actively contributing to the dataset's enhancement: The dataset enables the competition by providing training data, while the competition enriches the dataset with new unfolding attempts and diverse grasp strategies, creating a cycle of mutual improvement. This highlights the benefits of combining benchmarks with curated datasets to drive progress in robotic manipulation. 

In this paper, we present the benchmark for in-air robotic unfolding, the associated dataset and the outcomes of our cloth competition, which was held live at ICRA 2024 in Yokohama, Japan, as part of the 9th edition of the Robotic Grasping and Manipulation Competitions~\citep{yu2022rgmc, sun2024rgmc} (Figure~\ref{venue} provides an impression of the live trials).
We discuss insights gained into the current state-of-the-art in cloth unfolding, including the surprising effectiveness of hand-engineered, traditional geometric methods and the impact of the competition's design choices, such as the focus on single-attempt grasp selection. 
The winning team's promising 0.60 average coverage serves to ground other unfolding methods, but also underscores that robotic cloth unfolding remains an open challenge.

 In summary, this paper's three contributions are:
\begin{enumerate}
\item \textbf{A Novel Benchmark for Robotic Cloth Unfolding}: A standardized benchmark focused on objectively evaluating grasp selection strategies for unfolding garments.
\item \textbf{The ICRA 2024 Cloth Competition}: A unique head-to-head competition based on the proposed benchmark, involving 11 diverse teams and resulting in valuable insights into the state-of-the-art in robotic cloth manipulation.
\item \textbf{A Large Dataset of Cloth Unfolding Attempts}: A diverse dataset of 679 real-world robotic cloth unfolding attempts, used by the participants to prepare for the competition and extended with the data collected during the live evaluation trials of the participants' developed algorithms. 
\end{enumerate}
The dataset and benchmarking code are available at \url{https://airo.ugent.be/cloth_competition/}.


\section{Related Work} \label{sec:related-work}

\subsection{Cloth Unfolding} \label{sec:related-work-unfolding}

\subsubsection{Overview of Cloth Unfolding Strategies}

Cloth unfolding is a critical prerequisite for various downstream cloth manipulation tasks such as folding, hanging, ironing, and assisted dressing. 
Researchers have explored various unfolding strategies, with computer vision often playing an important role. 
For a general introduction to using computer vision for cloth manipulation, we refer the reader to the surveys by \citet{jimenez2017visual} and \citet{jimenez2020perception}.
The strategy evaluated in the benchmark and competition aligns with what Jimenez identifies as \textit{Grasp Point Localisation}, one of three key visual tasks he proposes as fundamental to cloth manipulation, alongside classification and state estimation.
Grasp point localisation is deemed fundamental because, as \citet{jimenez2017visual} aptly states: ``Any manipulative action requires instructing the robot where to grasp."
Combining well-placed grasps with open-loop motions can be sufficient for completing many manipulation tasks.
For instance, unfolding might be achieved with a predetermined stretch motion after determining two appropriate grasp poses on the cloth.

Jimenez further classifies grasp points into \textit{generic} and \textit{specific} categories.
Generic points, such as the highest and lowest points grasped when lifting crumpled garments, are typically not tied to specific features. On the other hand, specific points might target features like towel corners or sleeve cuffs. Table~9 in~\citet{jimenez2017visual}  provides a comprehensive overview of works categorized by targeted features and manipulation settings.
In the context of this competition, the term \textit{grasp pose selection} was preferred, emphasising the importance of gripper orientation and acknowledging that participants had the freedom to define semantic keypoints or employ alternative strategies. Whether targeting generic or specific points, or a combination thereof, the evaluation focused solely on the outcome: the degree of unfolding achieved after the grasp and stretch maneuver, approximated by the garment's surface area. This open-ended approach allowed for a wide range of solutions, including garment-category agnostic ones.

As an alternative to Jimenez's survey, a survey by \citet{borras2020grasping} focuses on the sequence of grasps and types of contact that are made and released during manipulation. The strategy benchmarked in this competition aligns with \textit{Task~1.~(a)~unfold~in~the~air} in Borras' Table~II.
To contextualize the competition and highlight the diversity in strategies, we present a brief overview of the different strategies explored in the literature on robotic cloth unfolding. 

\begin{itemize}
    \item \textbf{In-air grasping:} Grasping a new point in the air while holding the cloth with one gripper and then stretching the cloth to unfold~\citep{maitin2010folding, doumanoglou2014auto}. Starting from a crumpled garment, a common heuristic to bootstrap this strategy is first grasping its highest (or a random) point, lifting it, and then grasping its lowest point. This is the strategy used in the benchmark, and we also use the lowest point heuristic as the initialisation procedure.
    
    \item \textbf{Lifting and laying down:} Simultaneous grasping of two points on a cloth lying on a surface, followed by lifting and laying it back down~\citep{martinez2017recognition, thananjeyan2022luv}. FlingBot exemplifies this approach, incorporating a flinging motion before laying down the cloth~\citep{ha2022flingbot}. 
    
    \item \textbf{Edge tracing:} Sliding the cloth, preferably an edge, between the gripper's fingers until a corner or other desirable grasp for unfolding is reached. This has been tried without~\citep{shibata2010robotic, yuba2017unfolding} and with tactile sensing~\citep{salleh2008inchworm, proesmans2023unfoldir}. A core challenge of this strategy is grasping a good place to start tracing and then maintaining control of the grasp depth during tracing. This strategy is most straightforward for towels but has not been performed on a wide range of garments.
    
    \item \textbf{Dragging on a surface:} Grasping a point and pulling in a direction, utilising surface friction for unfolding~\citep{willimon2011model, sun2015accurate}. 
    This framing with top-down pick-and-place actions 
    has been popular for advanced learning-based techniques~\citep{hoque2022learning} such as reinforcement learning~\citep{lin2021softgym, mulero2023qdp}, imitation learning~\citep{seita2020deep, lee2024learning} and learning dense representations~\citep{ganapathi2021dense}.
    
    \item \textbf{Fold elimination:} Targeting and eliminating specific folds, usually after initial unfolding by other strategies~\citep{stria2017modelfree, ESTEVEZ2020103330, triantafyllou2022type}.
    
    \item \textbf{Brushing or sweeping:} Removing final wrinkles from nearly flattened cloth by moving the gripper along the surface without grasping, which is an example of non-prehensile cloth manipulation~\citep{paraschidis1995robotic, doumanoglou2016folding}.
\end{itemize}

While not mutually exclusive (humans often seamlessly switch between strategies), these categories offer a framework for understanding the diverse approaches in the literature. 
A key commonality among the first three categories is the focus on achieving two ``good" grasps on the cloth as a primary means of unfolding. These strategies operate under the generally valid assumption that grasping a cloth item at two suitable points can largely accomplish unfolding. Furthermore, due to the nature of the unfolding task, nearly all the aforementioned strategies can be applied iteratively, with the option to reset by simply releasing the cloth and starting over. This iterative nature necessitates that evaluation metrics like success rate or final coverage are always contextualized in the total amount of time taken, as even very inefficient strategies can eventually succeed through sufficient attempts.

\subsubsection{Unfolding Cloth through in-Air Regrasping}

This competition benchmarked the cloth unfolding strategy of unfolding in the air through regrasping.
This involves executing a well-placed grasp on a piece of cloth that is held in the air by one gripper.
This unfolding strategy has several compelling advantages, notably its simplicity and efficiency. This strategy often requires only two or three grasps to achieve complete unfolding, leveraging gravity to assist the process. 
Furthermore, it is compatible with standard parallel grippers, and it does not rely on the availability of large, flat surfaces, enabling its use in most environments.
The strategy also allows for iterative regrasping in the air, providing opportunities to recover quickly from suboptimal grasp selections.
The critical phase in the unfolding process for this strategy occurs when the cloth is held by a single gripper and drapes under the influence of gravity, as can be seen in Figure~\ref{hardware}.
This results in an intricate cloth configuration influenced by numerous factors, including garment shape, grasp location and depth, the number of grasped layers, the preceding lifting motion, material properties, friction, and buckling. This complex, occluded configuration makes it challenging to perceive the cloth's features and state accurately and to determine suitable grasp locations for unfolding.

Early efforts in robotic cloth unfolding began with Hamajima and Kakikura~\citep{hamajima1996planning}, who explored grasp point detection on garments suspended by a single gripper, targeting semantic keypoints such as the neck or sleeve tip. Their subsequent work shifted focus to grasping hemlines, utilising wide, shadowed regions near convex points in the cloth outline for detection~\citep{HAMAJIMA2000145}. However, recognizing that achieving two grasps on hemlines alone is often insufficient for full unfolding, Kaneko et al.~\citep{kaneko2001planning} developed an algorithm to identify the global configuration of cloth held by two grippers, by looking at the parts that stick out in a polygonal approximation of the cloth outline.
They later extended this to include the detection of ``adjacent characteristic parts," such as the left sleeve region and the left waist corner, enabling complete unfolding by placing the cloth on a table and grasping these points~\citep{Kaneko_2003jrm}. Osawa et al.~\citep{Osawa_2007jaciii} furthered this line of inquiry by analysing cloth held by two grippers with a novel approach of gradual stretching, utilising images of the cloth at intermediate stretched states for enhanced recognition. Additionally, they clearly described how iterative lowest-point grasping converges the grasp location to a discrete set of extremities. Despite these advancements, these early works relied on a final table-based fold elimination phase, rather than achieving complete unfolding in the air.

Hata et al.~\citep{hata2008robot} instead aimed at completing the unfolding entirely in the air by directly targeting a towel's corners adjacent to the held corner for grasping. Their method leveraged stereo depth and curvature analysis of both the 3D cloth boundary and the cloth surface for corner detection. Furthermore, they estimated the corner orientation using two points adjacent to the corner, extracted from the detected cloth boundary. \citep{maitin2010folding} also focused on in-air unfolding, employing depth and curvature information and convincingly demonstrated their towel unfolding on a PR2 robotic platform. 
Bersch et al.~\citep{bersch2011bimanual} explored grasping based on geodesic distance (distance along the cloth surface), using a shirt with markers. They highlighted the challenge in cloth manipulation arising from the discrepancy between Euclidean and geodesic distances: points that are close in Euclidean space can be distant parts of the cloth when considering the path along its surface.
This closeness can lead to the inadvertent grasping of multiple cloth parts and layers, hindering unfolding. Notably, they also pioneered the use of machine learning to predict the success of a grasp on hanging cloth, employing an autonomous data collection procedure to generate training data. Doumanoglou et al.~\citep{doumanoglou2014auto} further advanced the application of machine learning by using it for grasp location prediction, utilising predefined semantic keypoints (e.g., shirt shoulders) as target grasp locations. Their work was subsequently extended to predict suitable approach directions for grasping these corners~\citep{doumanoglou2014active} and integrated into a complete unfolding and folding pipeline~\citep{doumanoglou2016folding}.

Several works have continued exploring the use of supervized machine learning, primarily deep neural networks, to detect local features on hanging cloth.  While many studies have continued directly detecting semantic keypoints as grasp locations~\citep{CORONA2018629, saxena2019garment, zhang2020learning}, others have opted to detect other features such as cloth boundaries~\citep{twardon2015interaction, gabas2017physical}. This latter approach offers greater generality across cloth categories compared to category-specific keypoints, but grasping these boundaries may not be as effective for completely unfolding garments.  Ren et al. took the most comprehensive approach to local feature detection, recognizing both keypoints and larger semantic edge regions, such as necklines and cuffs~\citep{ren2023grasp}. However, effectively utilising these diverse features in a manipulation strategy remains a key challenge. This highlights a crucial consideration for perception-driven cloth manipulation: accurate perception alone does not guarantee successful grasping and task completion.

An alternative approach to local feature detection focuses on understanding the cloth's overall deformed global shape, also called pose or configuration.  Pioneering this direction, Kita et al.~\citep{kita2002model} developed a method for matching observations of real cloth with simulated models. This work, along with subsequent research~\citep{mariolis2014pose}, argues that the deformed shape of hanging cloth is primarily determined by the location that is grasped on the cloth. Garment pose recognition is treated as a discrete classification problem by discretising possible grasping points. Once a corresponding simulated model is matched, a predetermined grasp location can be selected.
Li et al.~\citep{li2015regrasping} proposed the most advanced version of this simulation-based matching. Their method involves matching against a large database of pre-simulated canonical references, applying rigid transformations (rotation, translation, scaling) using Iterative Closest Point, and performing non-rigid registration to improve the match between the simulated and observed cloth further. The predetermined grasp location from the matched simulation is then mapped to the reconstructed mesh of the real cloth and further refined based on local curvature to enhance graspability.

As the use of learning-based methods has increased in robotic cloth manipulation, so has the need for data. Collecting and annotating real-world cloth data is time-consuming and challenging to do accurately. Therefore, many learning-based techniques rely, at least partially, on synthetic depth maps~\citep{mariolis2014pose, CORONA2018629, ren2023grasp}, and recently also on high-fidelity rendered color images~\citep{lips2024learning}. Synthetic data offers a cost-effective way to obtain large quantities of precisely annotated data. However, it introduces a notorious ``sim-to-real" gap, a discrepancy that can diminish the accuracy of models trained on synthetic data when applied to real-world scenarios.


The diverse landscape of approaches, ranging from geometric analysis and simulation-based methods to deep learning techniques, highlights the complexity of robotic cloth unfolding. Despite decades of research, a definitive solution remains elusive, and the lack of standardized benchmarks makes it difficult to compare performance across different methods directly. This motivates the need for a rigorous, real-world competition like the one presented in this paper.

\subsection{Datasets for Robotic Cloth Manipulation}

Datasets like Dex-Net 2.0~\citep{mahler2017dexnet20} and GraspNet-1Billion~\citep{fang2020graspnet1billion} were instrumental in advancing learning-based grasping, primarily for pick-and-place tasks with rigid objects. These works relied on analytical measures to determine grasp stability, focusing on feasibility and force closure. However, grasp selection for cloth unfolding presents a more complex challenge.  The deformable nature of cloth makes it difficult to analytically determine grasp success, as the interaction between the gripper and the fabric can lead to unpredictable deformations. Additionally, the relative grasp location on the cloth significantly influences the unfolding outcome, making it crucial to consider not only grasp stability but also the subsequent manipulation trajectory. This distinction highlights the need for specialized cloth manipulation datasets.

Existing datasets for cloth manipulation vary significantly in their format and focus.  
Most datasets provide annotations of cloth features on static RGB or depth images.
For example, DeepFashion is a large-scale dataset with annotated cloth keypoints and categories, but primarily in the context of people wearing the garments~\citep{liu2016deepfashion}.
There are no images of cloth in a crumpled state or hanging from a single point, which are configurations that occur during manipulation. 
This limits its usability for robotic manipulation.
In the context of manipulation, feature-centric datasets have been proposed for category classification~\citep{gabas2016classification},  whole cloth segmentation~\citep{aragon2013glasgow}, cloth part segmentation~\citep{ramisa2014learning}, and keypoint detection~\citep{lips2024learning}.  Some datasets also incorporate tactile data, which can offer valuable insights, such as whether there is any cloth in the gripper or material properties, such as stiffness~\citep{kampouris2016multisensorial}. 
However, learning state representation is not enough to perform manipulation; the extracted features need to be interpreted and translated into manipulation actions, which often requires significant engineering efforts to achieve successful task execution.

A different approach is taken by datasets that record interactions for end-to-end learning, like those used in FlingBot~\citep{ha2022flingbot} and SpeedFolding~\citep{avigal2022speedfolding}. These datasets include perception data, manipulation actions (often formulated as primitives), and the resulting state of the cloth. This format enables direct learning of actions that progress towards task completion, bypassing the need to translate perceived features into actions. However, collecting such data can be more time-consuming. Learning from Demonstrations using teleoperated trajectory datasets is also gaining traction, with examples including simple tasks like towel pickup~\citep{shafiullah2023bringing}. However, we are unaware of publicly available datasets that offer demonstrations of cloth unfolding. It is also unclear at the moment how well policies trained on these types of datasets will generalize. Yet another option to collect interaction datasets is to record demonstrations of humans performing cloth unfolding~\citep{verleysen2020video}. Such demonstrations can be easier to collect. However, they introduce an "embodiment gap" due to the differences between human and robot capabilities and do not explicitly contain information about the actions taken by the humans.

The interaction datasets mentioned so far~\cite{ha2022flingbot,avigal2022speedfolding,verleysen2020video} focus on iterative table-based unfolding with simulated data or a single garment. In this work, on the other hand, we introduce an interaction dataset for in-air unfolding. The dataset contains interactions with 34 unique cloth items, utilizing a grasp-and-stretch action as motion primitive, for which grasp points are annotated. Efficiently collecting data for in-air cloth unfolding requires an informed grasp selection strategy, as random grasps are often ineffective. Therefore, we employed human-annotated grasp poses to bootstrap the dataset and ensure a sufficient number of successful unfolding attempts for the participants to learn from.

\section{Benchmark} \label{sec:benchmark}

\subsection{Task overview} \label{sec:benchmark-procedure}

\begin{figure*}[!t]
    \centering
    \includegraphics[width=\linewidth]{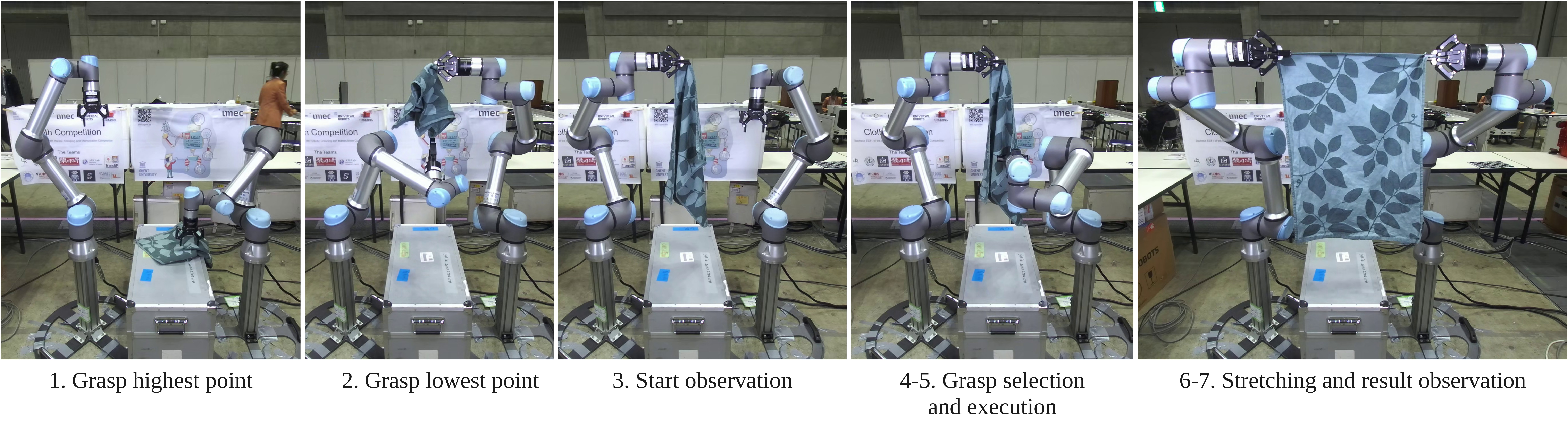}
    \caption{Overview of the unfolding procedure. After the robot grasps the cloth at its highest and then its lowest point, the grasp selection algorithm must select a grasp pose based on the RGB-D observation. The robot then executes this grasp and stretches the cloth, unfolding it in the process. A final observation is recorded for evaluation.}
    \label{steps}
\end{figure*}

The goal of this benchmark is to evaluate grasp algorithms for in-air robotic cloth unfolding. Robotic unfolding is the ability to untangle and spread out a cloth item from a (crumpled) starting configuration. 
Inspired by prior work~\citep{maitin2010folding, doumanoglou2014auto}, we propose a multi-step unfolding procedure, using the in-air unfolding strategy (see Section~\ref{sec:related-work-unfolding} for an overview). This procedure, which is illustrated in Figure~\ref{steps}, consists of the following steps:

\begin{enumerate}
    \item \textbf{Grasping the highest point:} The right arm autonomously grasps the highest point of the crumpled cloth on the table and lifts it.
    \item \textbf{Grasping the lowest point:} The left arm autonomously grasps the lowest point of the now-hanging cloth.
    \item \textbf{Start observation:} An RGB-D image of the hanging cloth is captured.
    \item \textbf{Grasp selection:} The grasp selection algorithm under evaluation (provided by the participants during a competition) determines the grasp pose for the right arm.
    \item \textbf{Grasp execution:} The right arm executes the provided grasp.
    \item \textbf{Cloth stretching:} Both arms stretch the cloth until a tension of \SI{2}{\newton} is achieved or until it is clear that the previous step failed to grasp the cloth.
    \item \textbf{Result observation:} An RGB-D image of the stretched cloth is captured to evaluate the unfolding performance.
\end{enumerate}

The entire procedure runs autonomously, starting with a randomly crumpled piece of cloth on a table. 
This emphasis on autonomous robot execution ensures a realistic starting point for participants, reflecting the actual output of robotic manipulation rather than a human-assisted setup.
The key design principle for the benchmark and accompanying competition was to reduce engineering overhead for the participants, thereby maximising research value.
To achieve this, we streamlined the unfolding pipeline, allowing users to focus their efforts on the most critical and challenging aspect: step 4, the final grasp selection. 
This approach bypasses the need for users to spend time on other parts of the unfolding system, such as camera calibration, inverse kinematics, and motion planning.
Instead, they can develop algorithms for grasp selection in cloth unfolding—a complex problem that demands an advanced understanding of image, cloth shape, and behaviour.

\subsection{Grasp Execution} 

To maximize both grasp success and workspace coverage while avoiding premature collisions, a two-part motion strategy is used for executing the provided grasps. This strategy consists of first moving the arm to a collision-free pregrasp pose, positioned a few centimeters behind the final grasp pose (i.e., the orientation is the same, but there is a small translation along the negative z-axis of the gripper, compared to the actual grasp frame). Then, from this pregrasp pose, the gripper approaches the final grasp pose linearly. This approach allows for potential contact with the cloth but strictly avoids collisions with the robot itself or the environment. Upon reaching the actual grasp pose, the gripper is closed.
Various distances were iteratively tested to determine a reachable and collision-free pregrasp pose. 

Users are allowed to submit multiple grasp poses to deal with grasps that are out of reach for the robot. We sequentially iterate over their proposals until a reachable grasp pose is identified.

For collision checking, the environment surrounding the robots is modelled statically, with the cloth obstacle represented by a convex hull calculated from point cloud data within a fixed 3D bounding box. 
Collision checking is performed using the Drake robotics toolbox~\citep{drake}. 
To move the robot to the desired poses, we use joint-space motion planning with  RRTConnect from OMPL~\citep{sucan2012the-open-motion-planning-library} and analytic
inverse kinematics for the UR5e based on~\citep{villalobos2022ik}.

\subsection{Hardware setup}

\begin{figure}[!t]
    \centering
    \includegraphics[width=\linewidth]{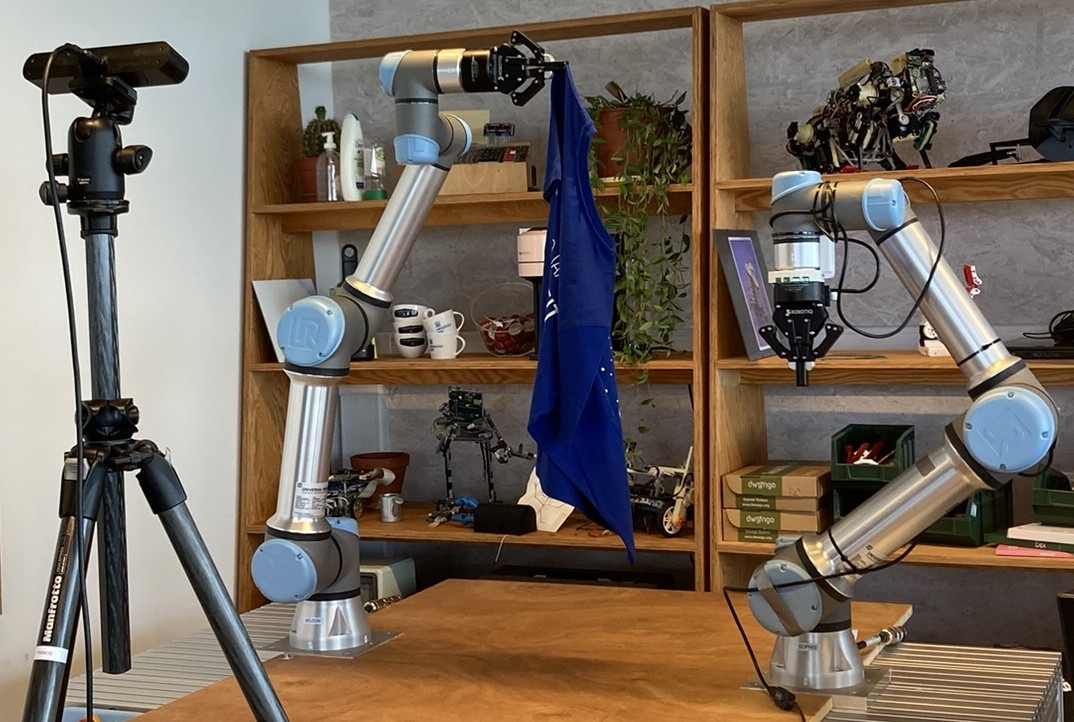}
    \caption{The hardware setup for the benchmark consists of a single RGB-D camera overlooking two UR5e robot arms. This image is from the AIRO lab setup, on which the training dataset was collected. Note that the wrist camera mounted on the right arm was not used for the competition.}
    \label{hardware}
\end{figure}

The hardware setup, of which an example is shown in Figure~\ref{hardware}, is designed to be easily reproducible and to mirror the anticipated capabilities of a general-purpose home assistant robot, approximating its workspace dimensions, reachability, and camera perspective.
A dual-arm setup is used to facilitate various cloth manipulation tasks and strategies. We use two UR5e collaborative robots, spaced \SI{90}{\centi\meter} apart. Both robots are equipped with internal wrist force-torque sensors and are widely used. Each arm is equipped with a Robotiq 2F-85 gripper featuring the standard rubber-coated fingertips. This parallel gripper, with an \SI{85}{\milli\meter} stroke, is also widely used and provides excellent pinch grasp strength, crucial for preventing cloth slippage, particularly during the final stretching motion.

A ZED 2i RGB stereo-depth camera is mounted \SI{80}{\centi\meter} above and \SI{140}{\centi\meter} behind the robots at a \ang{20} downward angle. This viewpoint emulates a head-mounted, ego-centric camera perspective of a household robot and should be reproduced as well as possible on different setups.

The ZED 2i camera has a focal length of \SI{2.1}{\milli\meter}, and the baseline distance between the two RGB camera lenses is \SI{12}{\centi\meter}. Depth maps are obtained with the ZED SDK (version 4.0), using the Neural depth mode. Algorithms should be provided with both RGB views in addition to the precomputed depth map and a point cloud, allowing them the flexibility to implement custom depth estimation or triangulation algorithms if desired.

Scene conditions, such as the tabletop on which the garments are initially placed, the background or the lighting conditions, are not explicitly specified, as we encourage the development of algorithms that are robust to these factors. Nonetheless, we recommend ensuring there is enough ambient light and adding additional light sources if required.

Other robot arms, grippers and cameras with similar properties could be used, although this would require changes in the accompanying codebase and might influence evaluation results.

\subsection{Evaluation} \label{sec:benchmark-evaluation}

\subsubsection{Measuring Performance}

\begin{figure*}[!t]
    \centering
    \includegraphics[width=\linewidth]{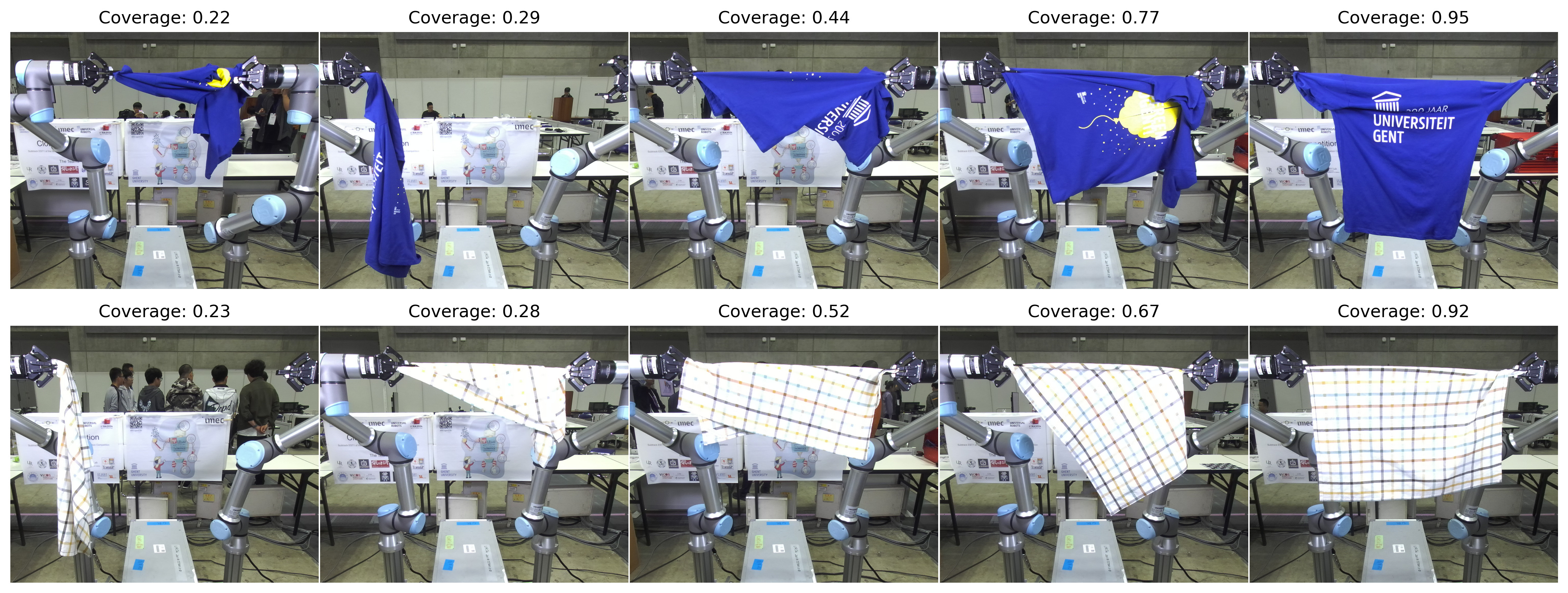}
    \caption{Examples from the competition of cloth unfolding results with varying coverage. Higher coverage generally corresponds to a more unfolded state, aligning with human assessment of task completion. Some grasps can be counterproductive, yielding lower coverage than not grasping at all.}
    \label{coverage}
\end{figure*}

For this benchmark, the quality of the grasp pose selection is assessed by the resulting degree of cloth unfolding, quantified by \textit{coverage} as in~\citep{seita2020deep, ha2022flingbot, hoque2022learning, proesmans2023unfoldir}.
Coverage is defined as the ratio between a garment's current and maximum potential projected surface area.
This metric is objective and generally correlates with the human perception of well-unfolded cloth, as illustrated in Figure~\ref{coverage} with several examples of different coverages.
The reference maximum surface area for each garment is determined by manually inserting the cloth into the robot grippers and performing the stretch under ideal conditions.

While coverage serves as the primary evaluation metric, we also analyze grasp success, which is defined as the presence of cloth between the gripper's fingertips after grasp execution. To further quantify their effectiveness and to provide insight into the failure modes of the grasp selection algorithms, we also introduce \textit{coverage for successful grasps}, which considers average coverage only for trials where grasping was successful.

The execution time of cloth manipulation algorithms is also an important metric to monitor~\cite{avigal2022speedfolding,irene2020benchmarking} as some methods require far more interactions or employ elaborate recovery schemes. For this benchmark, which restricts the interactions with the cloth to a single grasp and stretch, we recommend setting a reasonable maximum execution time (30 seconds seems reasonable) for the grasp selection algorithm. In addition, the average execution can be reported, but this is optional.

Evidently, as more unfolding trials are executed for each algorithm, the results become more significant. On the other hand, this also increases the burden on the evaluator. We recommend running at least 15 evaluations per algorithm, which is in line with other work on unfolding~\cite{avigal2022speedfolding,ha2022flingbot}. 

\subsubsection{Garments} \label{sec:benchmark-evaluation-garments}
We do not include a specific set of garments to test on for this benchmark, to avoid the additional effort of distributing such a physical dataset. We recommend ensuring diversity in terms of appearance, cloth materials and garment types, as cloth manipulation algorithms should be able to generalize to these properties. 
If possible, we recommend including towels from the Household Cloth Dataset~\cite{irene2022household}, which is a valuable effort to standardize cloth manipulation but lacks the visual diversity to properly test generalization. 
The garment set used for this competition can serve as inspiration.

For evaluating cloth manipulation, the initial configuration of the garment usually influences the algorithmic results significantly~\cite{gusseme2025insights}, warranting a protocol~\cite{proesmans2023unfoldir} to ensure the distribution of initial configurations is both appropriate and reproducible (as far as this is possible for clothes). However, this makes real-world evaluation even slower. In this benchmark, the initial configuration is less important as we use a two-phase initialization by grasping the highest and lowest point, as explained in Section~\ref{sec:benchmark-procedure}. As such, to reduce the evaluation effort, we recommend simply placing the cloth manually on the surface in a random configuration before initializing the evaluation procedure. Nonetheless, the initial configuration will remain a source of randomness. We believe this is unavoidable for cloth benchmarking, as it is infeasible to reproduce a garment configuration entirely. To limit the impact, evaluators should ensure they make the initial configurations as random as possible and as similar as possible between different evaluations.

\section{Dataset}

\subsection{Data Format}

\begin{table*}
\centering
\resizebox{\textwidth}{!}{%
\begin{tabular}{lllr}
\toprule
\textbf{File Name} & \textbf{Description} & \textbf{Shape \& Type} & \textbf{Approx. Size} \\ 
\midrule
\texttt{image\_left.png} & Left RGB image & 2208$\times$1242$\times$3 uint8 & 4 MB \\ 
\texttt{image\_right.png} & Right RGB image & 2208$\times$1242$\times$3 uint8 & 4 MB \\ 
\texttt{depth\_image.jpg} & Depth image, for visualization & 2208$\times$1242 uint8 & 200 kB \\ 
\texttt{depth\_map.tiff} & Depth map, aligned with left view & 2208$\times$1242 float32 & 10 MB \\ 
\texttt{confidence\_map.tiff} & Depth confidence map (from ZED SDK) & 2208$\times$1242 float32 & 10 MB \\ 
\texttt{point\_cloud.ply} & colored 3D point cloud, in world coordinates & 2742336 float32 & 40 MB \\ 
\texttt{arm\_left\_joints.json} & Left robot arm joint positions & 6 float32 & \textless{} 1 KB \\ 
\texttt{arm\_left\_pose\_in\_world.json} & Left robot arm base pose & 4$\times$4 float32 & \textless{} 1 KB \\ 
\texttt{arm\_left\_tcp\_pose\_in\_world.json} & Left gripper TCP pose & 4$\times$4 float32 & \textless{} 1 KB \\ 
\texttt{arm\_right\_joints.json} & Right robot arm joint positions & 6 float32 & \textless{} 1 KB \\ 
\texttt{arm\_right\_pose\_in\_world.json} & Right robot arm base pose & 4$\times$4 float32 & \textless{} 1 KB \\ 
\texttt{arm\_right\_tcp\_pose\_in\_world.json} & Right gripper TCP pose & 4$\times$4 float32 & \textless{} 1 KB \\ 
\texttt{camera\_intrinsics.json} & Camera intrinsic parameters & 3$\times$3 float32 & \textless{} 1 KB \\ 
\texttt{camera\_pose\_in\_world.json} & Camera extrinsic parameters & 4$\times$4 float32 & \textless{} 1 KB \\ 
\texttt{right\_camera\_pose\_in\_left\_camera.json} & Right camera pose relative to left camera & 4$\times$4 float32 & \textless{} 1 KB \\ 
\bottomrule
\end{tabular}%
}
\caption{Files contained in each observation of the dataset. Resolution and type are given for the resulting data after loading with our Python code. }
\label{tab:data_format}
\end{table*}

\begin{figure}[!t]
    \centering
    \includegraphics[width=\linewidth]{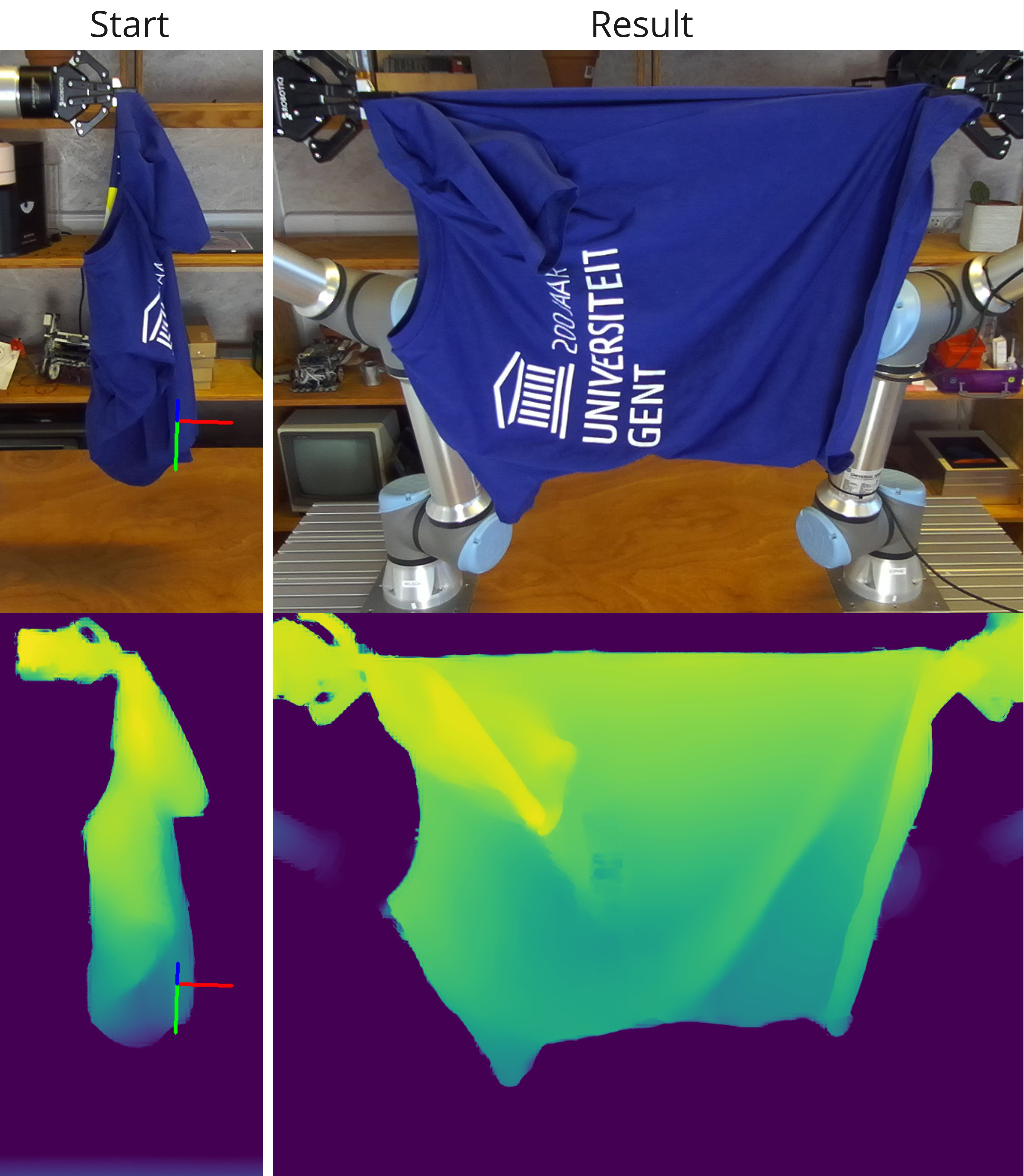}
    \caption{The color and depth images from the start and result observation from a dataset sample. The executed grasp is visualized on the start observation as an RGB coordinate frame representing the position and orientation of the gripper. The red axis indicates the direction along which the parallel gripper opens. By including the result observations, participants can estimate how effective the grasp was at unfolding.}
    \label{start_result}
\end{figure}

The ICRA 2024 Cloth Competition dataset focuses on grasp selection on hanging cloth for unfolding.
Each data sample documents a complete robotic unfolding attempt, consisting of a start observation, a grasp pose, and a result observation (Figure~\ref{start_result}).
The start observation captures the garment in a hanging state at the moment a grasp has to be selected. The grasp data includes the grasp pose selected for the robot. The result observation captures the garment's state after the grasp is executed and the robot arms have moved to stretch the cloth.
Each observation contains the stereo RGB images, depth and confidence maps, a colored point cloud, robot joint angles, gripper poses, and camera parameters.
Additionally, each episode includes a video recording of the entire unfolding attempt to provide enhanced context.

The dataset is organized using a straightforward folder structure. Each observation is contained within its own folder, and each unfolding interaction includes the initial observation, a grasp folder, the final observation, and a video file of the entire robot interaction.  
All data is stored in common and easily accessible formats. RGB images are stored as PNG files, while depth data is stored as both TIFF files for full precision and JPEG images for quick visualization. colored point clouds are stored in the PLY format. The video is encoded with a h265 codec and stored in an MP4 format.  All other data is stored in human-readable JSON files.
Details about the files stored in an observation are given in Table~\ref{tab:data_format}.
The world frame used as the main reference frame is positioned between the two robot bases, with robot arms mounted on the world Y-axis (the left arm on the positive side).

To facilitate easy access and analysis, the dataset's GitHub repository\footnote{\url{https://github.com/Victorlouisdg/cloth-competition}} includes Python code for loading data into NumPy arrays, along with several Jupyter notebooks. These notebooks demonstrate various functionalities, such as data visualization, depth-based background removal, conversion to a COCO keypoint dataset, and integration with popular Python libraries like OpenCV and Open3D. The notebooks also provide a detailed exploration of the data and folder structure, and tools to collect additional data.

In addition to unfolding attempts, the dataset includes a reference observation for each cloth item. This observation shows the cloth fully unfolded and held in the air, providing a baseline measurement of the maximum achievable surface area for use in evaluation metrics that measure relative coverage.

\subsection{Data Collection}
The dataset comprises two distinct parts.
The first part of the dataset was collected to help participants prepare for the ICRA 2024 Cloth Competition. This training dataset contains 503 demonstrations, each documenting the robot's attempt to unfold a cloth item using the procedure described in Section~\ref{sec:benchmark-procedure}, with human-annotated grasp poses. 

The annotators were PhD researchers from IDLab-AIRO, some specializing in robotic cloth manipulation, while others had expertise in unrelated areas. This diverse expertise resulted in varying degrees of success, exposing participants to both effective and ineffective grasps. We found it useful to provide the annotators with a virtual top-down view, obtained by reprojecting the point cloud, in addition to the actual camera viewpoint. They annotated the grasp position on the point cloud, and the orientation was chosen to be tangential to the point cloud. Although the annotators used various strategies, many used their internal understanding of clothes and tried to locate semantic locations on the cloth that were visible and appropriate for the other grasp point. This is similar to the methods used by earlier work~\cite{hata2008robot,maitin2010folding,doumanoglou2016folding}. Different annotators had different risk-reward trade-offs when choosing to grasp partially occluded semantic locations rather than locations that would result in lower coverage but were easier to grasp.

\begin{figure*}[!t]
    \centering
    \includegraphics[width=\linewidth]{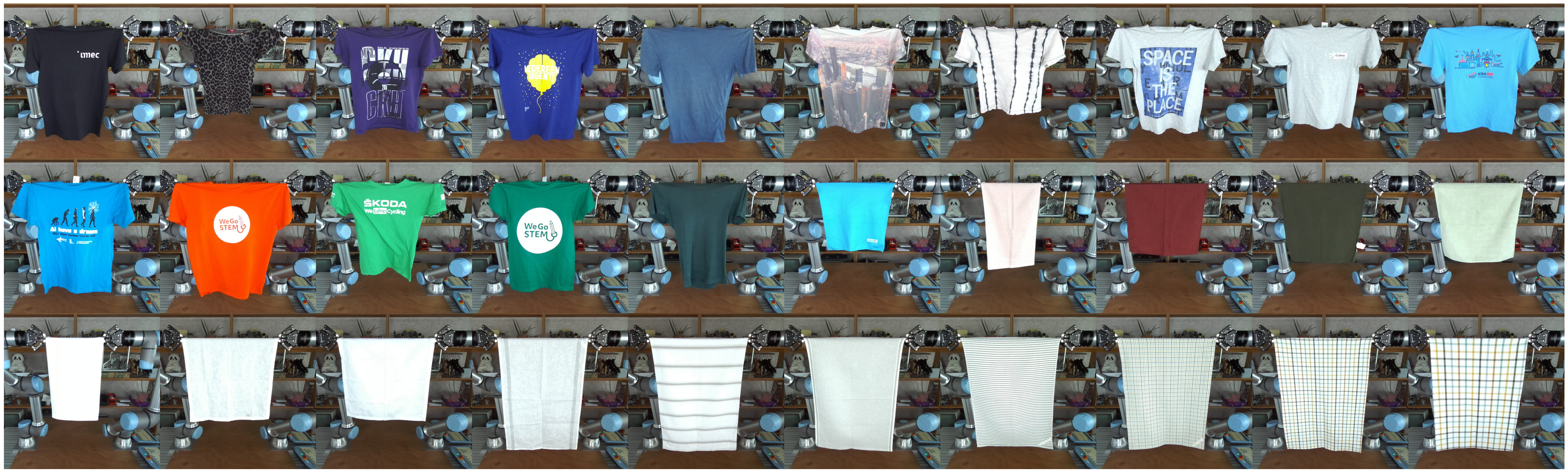}
    \caption{The 30 cloth items used for the training dataset with diverse colors, materials, sizes, and shapes. The 10 towels in the bottom row were selected from the Household Cloth Object Set~\citep{irene2022household}.}
    \label{training}
\end{figure*}

The training data familiarized participants with the competition setup, initialization procedure, grasp execution, stretch motion, and the diversity of cloth items. All training data was collected in a fixed lab environment, though the depth-based segmentation facilitates generalization to other environments. The 30 garments used during the collection of this training subset are shown in Figure~\ref{training}. They consist of 15 T-shirts and towels with varying shapes and appearances. 10 towels were selected from the Household Cloth Object Set~\cite{irene2022household}.

In addition to this training subset, all 176 evaluation trials conducted at the ICRA 2024 Cloth Competition were recorded and integrated into the dataset, expanding it to 679 episodes across 34 varied garments and enriching it with diverse grasp strategies and cloth items. The garments used during the competition are shown in Figure~\ref{evaluation} and are further discussed in Section~\ref{sec:competition-procedure}.
This resulted in the largest publicly available dataset of its kind, encompassing both human-annotated grasps and competition trial data, and enables in-depth analysis of algorithmic performance and failure modes. These features distinguish the dataset from prior work, which is often limited to a few real-world examples, and make it a valuable resource for future research in robotic cloth manipulation, primarily for developing unfolding systems. Although the dataset was collected using a single setup (dual-arm UR5e and ZED2i camera), we believe it can be used for arbitrary dual-arm setups, as it is straightforward to segment the cloth from the images and point clouds to remove the background and robot arms.



Note that some demonstrations in the dataset contain suboptimal grasp actions. The dataset (both train demonstrations and competition trials) includes trials with grasp failures (71/505) and/or low coverage. We did not remove these as they could be useful in some contexts, but we encourage users to filter the dataset to suit their needs. To facilitate filtering, the grasp success rate is available for each sample. Additionally, we provided a segmentation mask of the cloth in the final observation and a reference mask of the cloth item, which can be used to calculate coverage for each demonstration.

\section{Competition} \label{sec:competition}

\subsection{Procedure} \label{sec:competition-procedure}
During the ICRA 24 Cloth competition, we used the benchmark protocol described in section~\ref{sec:benchmark} to evaluate the grasping algorithms provided by the eleven participating teams.

\begin{figure}
    \centering
    \includegraphics[width=\linewidth]{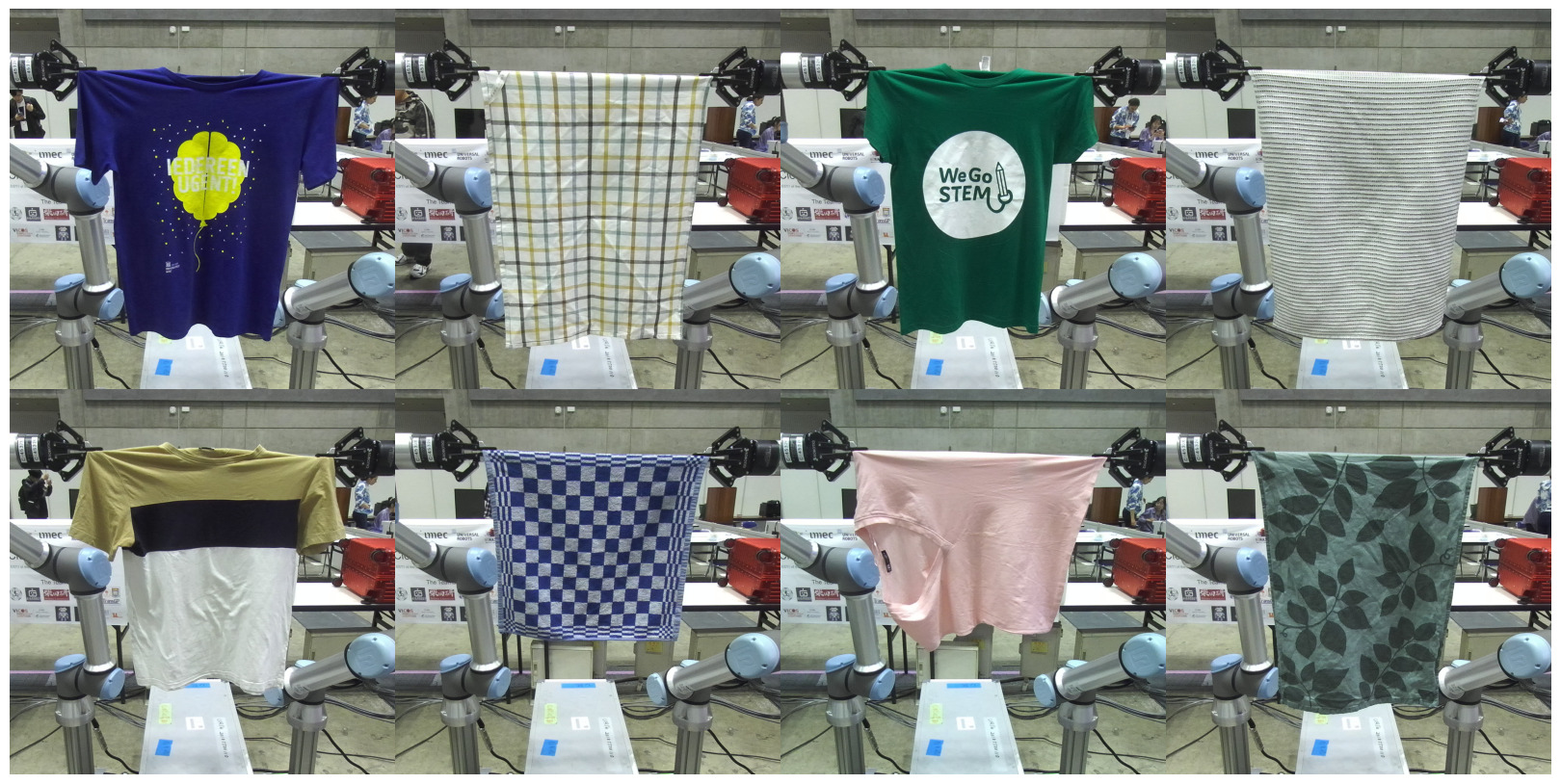}
    \caption{The cloth items used for the ICRA 2024 Cloth Competition. The items in the first row were also present in the training dataset, while the items in the second row were previously unseen items.}
    \label{evaluation}
\end{figure}

As discussed in Section~\ref{sec:benchmark-evaluation-garments}, the benchmark protocol does not impose a fixed set of garments due to the associated difficulties. For this competition, we selected eight garments to unfold. The garments are shown in Figure~\ref{evaluation} and include four T-shirts and four towels, with diverse sizes and appearances. Four of the eight garments were also included in the provided training dataset, whereas the others were unknown to the participants to avoid overfitting and to measure the generalization of their algorithms.

The teams had to unfold each garment twice, using the procedure described in Section~\ref{sec:benchmark-procedure}. The participants were allowed up to 30 seconds for their grasp selection algorithm for each unfolding attempt and were allocated a 50-minute timeslot to complete the 16 unfoldings.

The final score of each team is their average coverage across all trials. This includes the trials where the grasp participants selected led to grasp failure, leading the cloth to be held only by one of the grippers. Such scenarios yield low coverage, incentivizing participants to select secure grasps.

The evaluation for this competition was conducted live at ICRA 2024. This live setting ensured a rigorous assessment of grasp selection algorithms under real-world conditions, outside the participants' control and fostered interaction between the participants and members of the research community. 

Unlike competitions requiring teams to bring their own robot setups, participants were only required to bring  (or have access to) computing hardware to run their grasp selection algorithms.
The initial observation for each unfolding trial was downloaded over a local network, and they had to upload their chosen grasps to our server for execution. This created a truly plug-and-play competition environment, significantly lowering the barrier to entry.

\subsection{Participant's Grasp Selection Algorithms}
The participants of the ICRA 2024 Cloth Competition employed a wide array of approaches for grasp pose selection, which we categorize into two primary categories: \textit{traditional methods} and \textit{learning-based techniques}. 
Traditional methods rely on engineered image or point cloud processing techniques, while learning-based methods leverage machine learning, particularly deep neural networks.
Furthermore, we subcategorize the learning-based methods by their supervision scheme, highlighting the variety of learning paradigms used in the competition.

We classify two competition entries as \textit{traditional methods}. 
The first approach, \textit{Intuitive Grasping Determination} (IGD) by the AIR-jnu team, involves fitting a low-resolution polygon to the cloth's boundary after segmenting it from the RGB image. The second furthest vertex from the gripper holding the cloth is selected as the grasp point, and the grasp orientation is fixed.
This straightforward strategy is based on the observation that grasping the lowest is rarely the best choice. Consider a towel: the lowest point is usually the corner opposite the one held, while the ideal grasp points are the adjacent, higher corners. 
The intuition behind this approach is that the second furthest vertex, while not an optimal solution, offers a better chance for successful unfolding.

The second traditional method, \textit{Sharp Edge Detection} by EWHA Glab, builds on prior research on identifying sharp points in point clouds~\citep{weber2010sharp}. These prominent points are considered likely candidates for successful grasping. Subsequently, the candidate closest to the camera is designated as the grasp location. The grasp orientation is aligned with the point cloud normals, except for edge points, where the direction is aligned towards the mean point of a small neighbouring point cluster.
This method prioritized graspability through careful consideration of both location and gripper orientation.

The remaining nine methods are learning-based methods, albeit with varying learning objectives. These methods span a spectrum from less to more task-informed. Some prioritize grasp success, while others emphasize the effectiveness of grasps for the specific task of unfolding. More specifically, four distinct supervision schemes can be distinguished: \textit{(1) Graspability Estimation}, \textit{(2) Semantic Keypoint Detection}, \textit{(3) Grasp Imitation}, and \textit{(4) Coverage Prediction}.

\textbf{(1) Graspability Estimation} focuses primarily on ensuring a successful grasp, which shares similarities with grasp prediction techniques used for rigid object pick-and-place tasks. In these scenarios, any stable grasp is considered adequate. Two teams within the competition adopted methods that specifically prioritize graspability.

The \textit{Affordance Edge Detection} (AED) method by UOS-Robotics employs a hybrid approach, combining traditional image processing and deep learning. It combines a canny edge detection on the depth map and identifies \textit{affordance regions} from RGB images using a semantic segmentation model trained on successful grasps. Overlapping these regions creates \textit{grasp regions}, which are mapped to the point cloud to identify clusters of high-affordance points on cloth edges as potential grasp candidates. In addition, the two largest triangles formed within the cloth, using the furthest points from the line connecting the TCP point and the lowest point, are identified. The two furthest points and the lowest point are also scored based on affordance and added to the candidate pool. The candidate with the highest affordance is ultimately selected as the grasp location. Finally, the robot's grasping direction is calculated using the least squares line through the grasp point. The team manually filtered the provided training data to remove demonstrations in which the robot and cloth were entangled.

The second learning-based method that focuses on graspability, \textit{Grasp-Cloth} by Samsung Research China - Beijing, starts by dividing the point cloud into key areas, for example, around the bottom left corner and the bottom right corner for towels. These key areas were validated as good grasp regions by using the PyFleX cloth simulator. Subsequently, grasps for each region are generated using a grasp generation network that they trained on Graspnet-1Billion~\citep{fang2020graspnet}, even though this dataset notably lacks cloth-like objects. This network is thus used purely to assess graspability in a geometric sense and has no notion of cloth behavior or the task of unfolding. 
This is also the only method that did not use a fixed grasp depth but instead used the grasp depth outputted from the neural network.

\textbf{(2) Semantic Keypoint Detection} aims to detect semantically meaningful points on the cloth, such as a shirt's left shoulder and right waist, or the corners of a towel. These keypoints can be considered inherent to the cloth items themselves, and their relevance extends beyond any single task. The \textit{KeypointDetr} method by SCUT-ROBOT explicitly implements semantic keypoint detection and is the only team that incorporates cloth category classification, distinguishing between towels and shirts~\citep{carion2020end}. However, detecting semantic keypoints alone is not sufficient for unfolding, additional grasp selection rules are necessary. For towels, grasping corners adjacent to the one held by the robot yields the best results. For shirts, the held semantic keypoint is determined, dictating a preferred sequence of grasp points (e.g., if the left shoulder is held, the order is right shoulder, left waist, right collar, right waist, left collar).

\bgroup
\def\arraystretch{1.3}
\begin{table*}[htbp]
    \centering
    \caption{Results of the competition, showing the average score for each metric across 16 evaluation trials. Methods are ranked based on their average coverage.}
    \label{results}
\begin{tabular}{cllccc}
\hline
\textbf{Rank} & \textbf{Team} & \textbf{Method} & \makecell{\textbf{Grasp}\\\textbf{success rate}} & \makecell{\textbf{Coverage for}\\\textbf{successful grasps}} & \textbf{Coverage} \\
\hline
1 & AIR-jnu & Intuitive Grasping Determination & {\cellcolor[HTML]{D1D2E7}} \color[HTML]{000000} 0.69 & {\cellcolor[HTML]{C1D9ED}} \color[HTML]{000000} 0.73 & {\cellcolor[HTML]{98D594}} \color[HTML]{000000} 0.60 \\
\arrayrulecolor{gray}\hline\arrayrulecolor{black}
2 & Team Ljubljana & CeDiRNet-6DoF & {\cellcolor[HTML]{D8D8EA}} \color[HTML]{000000} 0.63 & {\cellcolor[HTML]{C9DDF0}} \color[HTML]{000000} 0.69 & {\cellcolor[HTML]{A8DCA2}} \color[HTML]{000000} 0.57 \\
\arrayrulecolor{gray}\hline\arrayrulecolor{black}
3 & Ewha Glab & Sharp Edge Detection & {\cellcolor[HTML]{B6B6D8}} \color[HTML]{000000} 0.94 & {\cellcolor[HTML]{DAE8F6}} \color[HTML]{000000} 0.57 & {\cellcolor[HTML]{B1E0AB}} \color[HTML]{000000} 0.55 \\
\arrayrulecolor{gray}\hline\arrayrulecolor{black}
4 & SCUT-ROBOT & KeypointDetr & {\cellcolor[HTML]{BDBEDC}} \color[HTML]{000000} 0.88 & {\cellcolor[HTML]{D9E7F5}} \color[HTML]{000000} 0.58 & {\cellcolor[HTML]{BBE4B4}} \color[HTML]{000000} 0.53 \\
\arrayrulecolor{gray}\hline\arrayrulecolor{black}
5 & Team Greater Bay & CopGNN & {\cellcolor[HTML]{CBCBE3}} \color[HTML]{000000} 0.75 & {\cellcolor[HTML]{D6E5F4}} \color[HTML]{000000} 0.60 & {\cellcolor[HTML]{BBE4B4}} \color[HTML]{000000} 0.53 \\
\arrayrulecolor{gray}\hline\arrayrulecolor{black}
6 & Samsung Research China & Grasp-Cloth & {\cellcolor[HTML]{D8D8EA}} \color[HTML]{000000} 0.63 & {\cellcolor[HTML]{D7E6F5}} \color[HTML]{000000} 0.59 & {\cellcolor[HTML]{D1EDCB}} \color[HTML]{000000} 0.48 \\
\arrayrulecolor{gray}\hline\arrayrulecolor{black}
7 & Shibata Lab & Densenet Method & {\cellcolor[HTML]{DEDDED}} \color[HTML]{000000} 0.56 & {\cellcolor[HTML]{DAE8F6}} \color[HTML]{000000} 0.57 & {\cellcolor[HTML]{D9F0D3}} \color[HTML]{000000} 0.46 \\
\arrayrulecolor{gray}\hline\arrayrulecolor{black}
8 & AI\&ROBOT LAB & CFAN & {\cellcolor[HTML]{D8D8EA}} \color[HTML]{000000} 0.63 & {\cellcolor[HTML]{DAE8F6}} \color[HTML]{000000} 0.57 & {\cellcolor[HTML]{DCF2D7}} \color[HTML]{000000} 0.45 \\
\arrayrulecolor{gray}\hline\arrayrulecolor{black}
9 & UOS-Robotics & Affordance Edge Detection & {\cellcolor[HTML]{F6F5F9}} \color[HTML]{000000} 0.19 & {\cellcolor[HTML]{BCD7EB}} \color[HTML]{000000} 0.75 & {\cellcolor[HTML]{EEF8EA}} \color[HTML]{000000} 0.39 \\
\arrayrulecolor{gray}\hline\arrayrulecolor{black}
10 & AIS Shinshu & Depth2Grasp-CNN & {\cellcolor[HTML]{BDBEDC}} \color[HTML]{000000} 0.88 & {\cellcolor[HTML]{F7FBFF}} \color[HTML]{000000} 0.37 & {\cellcolor[HTML]{F2FAF0}} \color[HTML]{000000} 0.37 \\
\arrayrulecolor{gray}\hline\arrayrulecolor{black}
11 & 3C1S & Grasp with PointNet-VAE & {\cellcolor[HTML]{FCFBFD}} \color[HTML]{000000} 0.06 & {\cellcolor[HTML]{94C4DF}} \color[HTML]{000000} 0.91 & {\cellcolor[HTML]{F7FCF5}} \color[HTML]{000000} 0.35 \\
\hline
\end{tabular}
\end{table*}
\egroup

\textbf{(3) Grasp Imitation} methods train neural networks to imitate the grasp locations chosen during data collection, for example by human annotators. These grasps are inherently task-informed, as they are selected with the explicit intention of unfolding the cloth.

The \textit{Densenet Method} by Shibata Lab employs a neural network to regress the annotated grasp position from RGB images directly. It complements this position prediction with a fixed grasp orientation.

\textit{Depth2Grasp-CNN} by AIS-Shinshu also regresses grasp position but utilizes depth maps as input. The grasp direction is then determined through principal component analysis on the points within a \SI{10}{\centi\meter} radius of the predicted grasp point. The team augmented their training data with synthetic depth maps generated from simulations to enhance their model's performance.

The \textit{CFAN (Cloth Flattening with Affordance learning and surface Normal estimation)} method by AI\&ROBOT LAB employs a neural network to generate an affordance map from depth images~\citep{ sunil2023visuotactile}. The point with the highest affordance is selected as the grasp location, and the surface normal at that point determines the grasp orientation.
The \textit{Grasp with PointNet-VAE} method by 3C1S utilizes a PointNet-like architecture that takes the segmented colored point clouds and outputs two points.
The first point is used as grasp location and the second point, which has to be adjacent to the first, is used to determine grasp orientation.
This pose prediction is then refined further using a variational autoencoder (VAE).
The team manually filtered the provided dataset to remove data points with failed grasps or suboptimal coverage.

The \textit{CeDirNet-6DoF} method employed by Team Ljubljana utilises a neural network for direct regression of annotated position and orientation on RGB images. Building upon CeDiRNet~\citep{TABERNIK2024110540}, this method predicts center direction vectors, parametrized using a trigonometric function, where each pixel's vector points towards the nearest grasp point within an image. The team extended the CeDiRNet architecture by incorporating a novel 6-DoF prediction module, enabling the dense regression of Euler angles representing pose orientation. The model was pre-trained on a dataset of 6000 towel images, collected by the team, featuring corner annotations and 2D approach directions~\citep{TabernikRAL2024}. This pretraining phase, focusing on predicting towel corners, shares similarities with the semantic keypoint detection approach. Subsequently, it was fine-tuned on the 500 competition dataset examples, incorporating 3D approach directions. Notably, this method derives both grasp locations and orientations entirely from data, without relying on geometric rules.

\textbf{(4) Coverage Prediction} methods predict the cloth coverage achieved by a given grasp, making them highly task-relevant.
The \textit{Coverage Prediction GNN} method by Team Greater Bay utilises two graph neural networks: one predicts the physical connectivity between points in the cloth, while the second estimates the resulting coverage given this connectivity and a chosen grasp point~\citep{pmlr-v164-lin22a, yang2024one}. To achieve this, they first construct edges between nearby points in the point cloud and then employ an Edge GNN to determine which of these edges represent actual physical connections within the cloth. Once the cloth's structure is understood, a second GNN predicts the coverage area that would result from grasping at a particular point.
During deployment, candidate grasp points are selected based on heuristics to avoid challenging areas, such as ``valleys" where the cloth bends away from the gripper. The coverage prediction model then evaluates each candidate, and the point with the highest predicted coverage is chosen for grasping.

\subsection{Competition Results}

The live evaluations at ICRA 2024 showcased varying performance of the 11 competing methods.
The final ranking was determined by average coverage across 16 trials per team (two trials per cloth item) and ranged from 0.35 to 0.6, as can be seen in Table~\ref{results}.
Next to average coverage, we also report the overall grasp success rate and the average coverage of the successful grasps.
The latter indicates the effectiveness of the grasp in unfolding the cloth, for those attempts where the grasp succeeded, and allows for further analysis of limitations and failure modes.


Interestingly, the top two methods only achieved an average grasp success rate but excelled due to the high coverage of their successful grasp attempts, which is around 0.7.
This is significantly higher than all other methods (excluding those with a very low grasp success rate, where coverage for successful grasps might be unrepresentative).
Teams ranked between 3rd and 8th place displayed similar coverage given successful grasps of around 0.6, with variations mainly in grasp success rate.
Notably, the third-place team was able to secure their place by prioritising grasp success, failing to grasp only once. 

Two methods struggled to achieve grasp success, the \textit{AED} and the \textit{Grasp with PointNet-VAE} methods.
Despite its lower grasp success rate, the former approach showed promise but suffered from many grasps that failed by only a small margin, whereas the successful grasps were highly successful.
The latter method encountered technical difficulties, which prevented it from performing grasps until these were resolved on the ninth evaluation trial.

\begin{figure*}[!t]
    \centering
    \includegraphics[width=\linewidth]{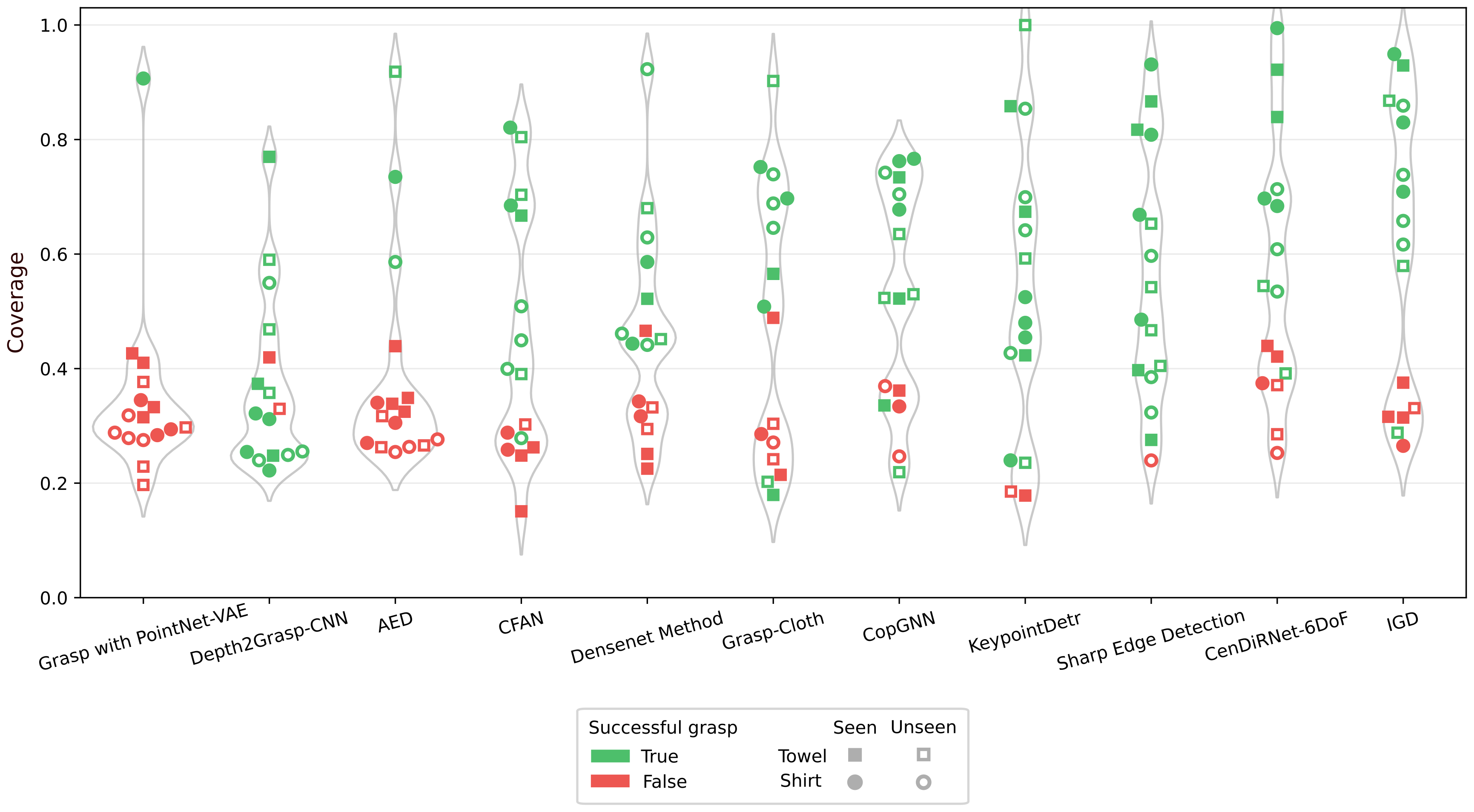}
    \caption{Coverage distribution for each method across 16 trials, illustrating the difference between methods, but also the large spread for each method individually.
    Methods are ordered left to right by increasing average coverage. Each marker represents an unfolding trial, with squares representing towels and circles for shirts. Hollow markers signify items not present in the training data. The green markers show that the cloth was successfully grasped. The plot reveals that while grasp success is crucial, the quality of successful grasps is the key differentiator.}
    \label{violin}
\end{figure*}

An even more detailed view of all 176 evaluation trial run at ICRA is given in Figure~\ref{violin}.
The analysis of all trials across all teams reveals several notable observations.
Shirts had a significantly higher grasp success rate (0.7) than towels (0.53), likely attributable to their lower bending stiffness, which facilitates grasping.  Despite this difference in grasp success, the average coverage achieved by successful grasps was similar for both shirts (0.60) and towels (0.57).  Furthermore, no significant difference in average coverage was observed between seen (0.49) and unseen items (0.47), suggesting that most methods generalize well to unseen garments.

Failing to grasp the cloth in most cases leads to significantly lower coverage, as can be seen in Figure~\ref{violin}. The variability in coverage for these failed trials stems from the randomness in initial configurations. Even though we grasp the highest and lowest point before starting the unfolding, the initial configurations of the garments as they lie on the table influence the configuration achieved before the grasp point is chosen by the participants. This randomness is unavoidable, as it is infeasible to reproduce arbitrary cloth configurations. However, its impact is reduced by the initialization procedure. Furthermore, despite this randomness, the coverage for failed grasps is, in general, significantly lower than trials with successful grasps. An unfortunate exception to this rule is the \textit{Depth2Grasp-CNN} method, which has many successful grasps that resulted in low coverage. This is because the method produced grasps with relatively central and deep placement within the cloth, often grasping multiple layers of fabric, which hindered the unfolding. This illustrates the importance of finding appropriate, goal-oriented grasps for unfolding, as mere grasp success is not sufficient.

\section{Discussion}

The competition successfully achieved its goal of establishing a realistic benchmark for out-of-the-lab in-air robotic cloth unfolding, addressing a pressing need within the robotics community.
In doing so, it also extended the competition dataset with valuable data collected using the participants' diverse grasp selection strategies.  

When comparing the competition results to prior work, a noticeable gap emerges between the average coverage achieved here (0.6 by the top team) and values reported in previous studies (e.g., 0.8 by FlingBot~\citep{ha2022flingbot} and 0.85 by UnFoldIR~\citep{proesmans2023unfoldir}). However, this discrepancy can be attributed to several factors that make direct comparisons challenging. Subtle differences in experimental setups, such as the cloth's initial configuration or the choice of evaluation items, can significantly influence unfolding difficulty. Moreover, prior work often reports coverage after multiple actions, recovery attempts, or filtering out grasp failures—all of which can substantially inflate performance metrics. The competition's live, single-attempt format, diversity in garments and automated evaluation minimize potential biases, offering a more realistic assessment of cloth unfolding capabilities under real-world constraints. This highlights the critical importance of independent evaluations outside controlled laboratory settings.

Shifting our focus to the performance within the competition itself, the broad range in coverage scores, which range from 0.35 to 0.6, highlights the significant progress made in cloth unfolding. This advancement is particularly striking when comparing the lowest coverage given successful grasps of 0.37 to the coverage of 0.73 achieved by the winning method. While there remains room for improvement before reaching the theoretical maximum coverage of 1.0, the top score is especially remarkable given the challenges posed by imperfections in the initialization procedure and the unsolvable occlusions inherent in the competition's single-attempt format. The 69\% grasp success rate of the winning method also suggests there are still opportunities to enhance performance.

\begin{figure}[!t]
    \centering
    \includegraphics[width=\linewidth]{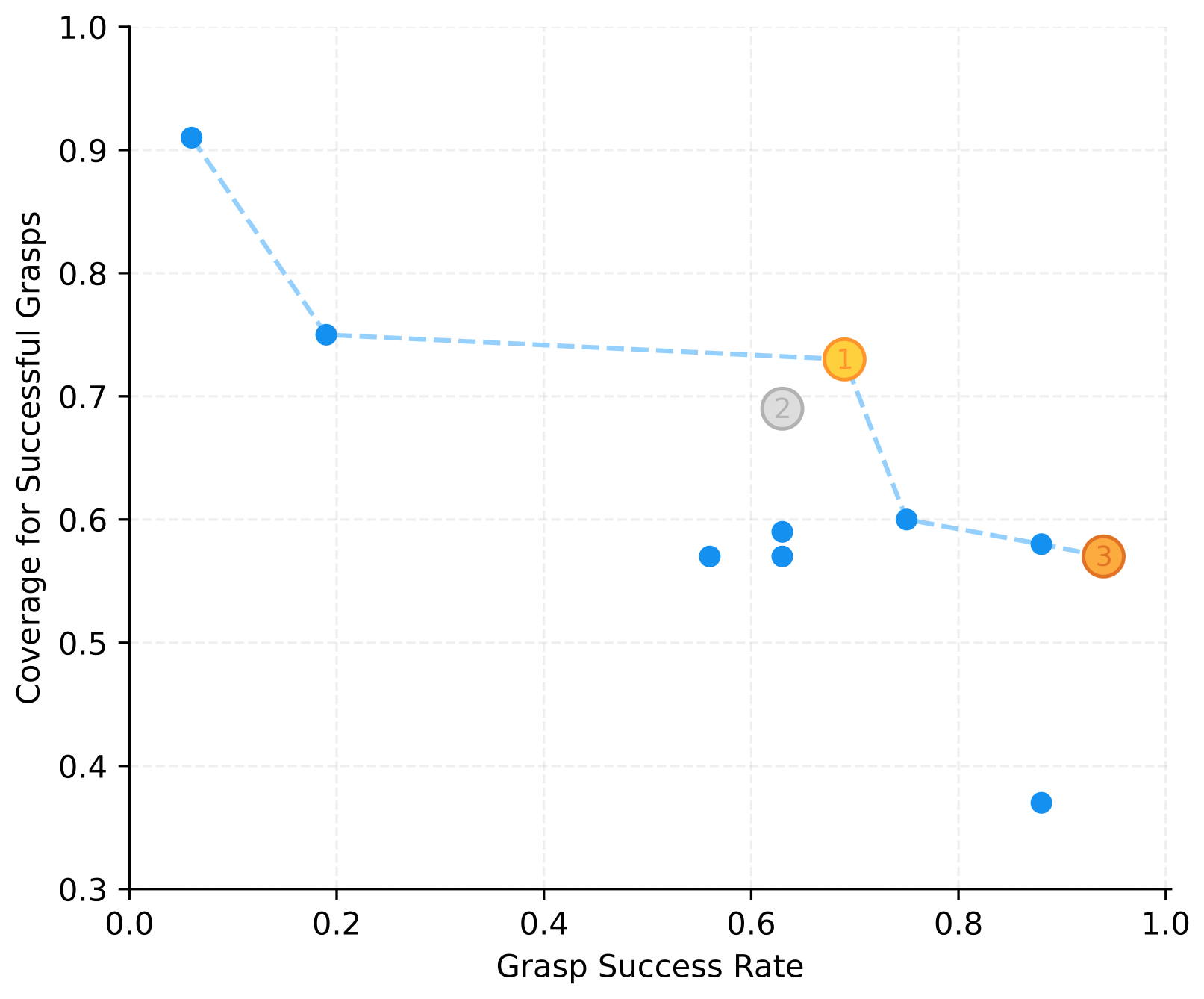}
    \caption{Illustration of the coverage for successful grasp and grasp success rate for each method. The three highest-ranked methods are shown as medals. The Pareto front is drawn as a dashed line. The plot shows a risk-reward trade-off, where strategies that aim for higher potential coverage might risk a higher chance of grasp failure.}
    \label{pareto}
\end{figure}

Further analysis reveals a risk-reward trade-off between grasp success and coverage, as illustrated in Figure~\ref{pareto}. Some teams prioritized reliable grasping, leading to high grasp success rates but potentially lower coverage. In contrast, others pursued greater coverage, accepting a potentially increased risk of grasp failure.
While average coverage is an objective metric that aligns reasonably with unfolding progress, it might have encouraged conservative strategies because the penalty for missing a grasp is large. As we have shown, the distribution of coverage values offers additional insights. Future competitions could explore more nuanced evaluation metrics, potentially rewarding attempts that achieve particularly high coverage (e.g., exceeding 0.8 or 0.9) even more.

The popularity of learning-based methods in robotic manipulation makes the strong performance of the two traditional, non-learning methods, securing first and third place, even more remarkable.
These geometric methods, which lack a semantic understanding of the grasped cloth part, excelled in both the coverage of their successful grasps and their overall grasp success rate.
This puts the performance of learning-based methods into perspective and shows that while learning-based methods are promising to achieve robustness and generalization, a lot can still be learned from more traditional methods.
The absence of semantic understanding in these geometric methods motivates further exploration of learning-based approaches that can exploit such understanding for improved performance.

The competitive performance of the second-placed method showcases the potential of learning-based approaches. It also illustrates the usefulness of the dataset for learning in-air unfolding and shows that such methods are able to generalize to different scenes and even unseen clothes, even when they use RGB images instead of geometric depth images as input. However, reaching their full potential likely requires even more high-quality data, which is expensive to collect on real robots. The competitiveness of traditional methods suggests they could play a key role in autonomous data collection, effectively serving to ``jump-start the data flywheel". 

An important dataset design choice was the exclusion of explicit class labels (towel vs. shirt) and semantic keypoint annotations (e.g., left-waist). By not providing such labels, we wanted to stimulate the development of more generic methods. While participants were free to annotate the competition dataset themselves, we wanted to break from previous work, where the cloth category is sometimes assumed to be known in advance, making cloth classification a potential failure point in the grasp selection pipeline. 
Only one team explicitly utilized cloth category classification and semantic keypoint detection, a surprising contrast to the emphasis on such techniques in the literature~\citep{doumanoglou2016folding, stria2018classification, CORONA2018629, saxena2019garment}. 
All other methods were class-agnostic, which could facilitate their extension to other cloth categories~\citep{triantafyllou2022type}. This is crucial given the real-world diversity of cloth items, not all of which can be easily categorized or annotated with consistent keypoints.

We are convinced that the competition dataset, now enriched with all evaluation trials, holds immense value for advancing learning-based cloth unfolding. It offers realistic starting configurations and outcomes of various grasp selection strategies, showcasing both successful and less effective approaches. It can hence be used for further improvement of grasp selection algorithms for in-air unfolding, but it could also be used for various other approaches that learn from demonstrations.
Despite the inherent cost of collecting such data, we encourage future competitions, particularly algorithm-level robotics benchmarks, to incorporate some robot execution data. 
This inclusion would not only serve to ground algorithms in the practical realities of cloth manipulation but would also be a valuable resource to participants, providing common reference point and fostering a shared understanding of the task's complexities.

A final note is the choice of an algorithm-level benchmark~\citep{yu2022rgmc} with a focus on grasp selection, contrasting with previous system-level cloth manipulation benchmarks~\citep{garcia2022clothcomp}.
Our targeted approach inherently directed participant efforts, giving a clear picture of the state of this strategy to cloth unfolding. 
Future iterations could explore granting more control, potentially allowing for grasp selection in initialisation, multiple regrasps, or even progress towards closed-loop control of the dual-robot arm and allowing participants to select their own unfolding strategy.
This could enable participation of a wider range of approaches (e.g., reinforcement learning~\citep{lin2021softgym, ha2022flingbot}, learning from demonstrations~\citep{zhao2024aloha, verleysen2020video}), though at the cost of increased operational and evaluation complexity.

\section{Conclusion}

The ICRA 2024 Cloth Competition successfully established a rigorous benchmark for in-air robotic unfolding, measuring performance outside of controlled laboratory environments on a shared setup using an objective metric. The strong participation and engagement demonstrated a clear desire within the community for standardized benchmarking in this domain and a strong appreciation for the chosen scope, task setup, and evaluation metrics. 
This underscores the effectiveness of the competition design in capturing the core challenge of perceiving and grasping hanging cloth and providing a meaningful platform for evaluating progress.

The competition results highlight a discrepancy between performance observed here and values reported in prior work, underscoring the critical importance of independent evaluations outside controlled laboratory settings.
The wide coverage distributions and the 0.60 average coverage of the winning team highlight that there is considerable room for improvement and that robotic cloth unfolding remains an open problem.
This is further emphasized by the diversity of competitive approaches, where, surprisingly, traditional methods outperformed learning-based approaches despite the latter's increasing popularity. High-quality robotic cloth manipulation data is crucial to drive progress and advance learning-based techniques in this domain. Recognizing this need, the competition yielded a comprehensive and publicly released dataset of 679 real-world cloth unfolding attempts.
This dataset, encompassing diverse manipulation strategies across various garments, is a valuable resource for enabling the development of novel grasp selection algorithms.

This competition, with its associated dataset, will hopefully stimulate the development of more benchmarks and datasets for cloth manipulation. Through iterative refinements and continued community engagement, this competition can evolve into a definitive benchmark for cloth unfolding, fostering further progress and collaboration in this challenging field.  
 
\section*{Authors' Affiliations}

\footnotesize Victor-Louis De Gusseme, Thomas Lips, Remko Proesmans, Andreas Verleysen, and Francis wyffels are with the AI and Robotics Lab (IDLab-AIRO), Ghent University - imec, Ghent, Belgium, supported by Research Foundation Flanders (FWO) under Grant Numbers 1SD4421N, 1S15923N, 1S56022N and euROBIn Project (EU grant number 101070596).

\footnotesize Julius Hietala is an independent researcher

\footnotesize Giwan Lee, Jiyoung Choi, Jeongil Choi, Geon Kim, and Phayuth Yonrith are with the Department of Mechanical Engineering, Chonnam National University, Gwangju, South Korea.

\footnotesize Domen Tabernik, Matej Urbas, Jon Muhovič, and Danijel Skočaj are with the Faculty of Computer and Information Science, University of Ljubljana, Ljubljana, Slovenia.

\footnotesize Andrej Gams, Peter Nimac, and Matija Mavsar are with the Jožef Stefan Institute, Ljubljana, Slovenia.

\footnotesize Hyojeong Yu, Minseo Kwon, and Young J. Kim are with the Ewha Computer Graphics Lab, Ewha Womans University, Seoul, Republic of Korea, supported by the ITRC project: “IITP-2024-2020-0-01460”, and were advised by Young J. J. Kim.

\footnotesize Yang Cong, Supeng Diao, Jiawei Weng, and Jiayue Liu are with the South China University of Technology, Guangzhou, China.

\footnotesize Ronghan Chen and Yu Ren are with the Shenyang Institute of Automation, Shenyang, China.

\footnotesize Haoran Sun is with the Southern University of Science and Technology, Shenzhen, China, and The University of Hong Kong, Hong Kong SAR.

\footnotesize Linhan Yang is with the Southern University of Science and Technology, Shenzhen, China, The University of Hong Kong, Hong Kong SAR, and Center for Transformative Garment Production, Hong Kong SAR.

\footnotesize Zeqing Zhang, Lei Yang, and Jia Pan are with The University of Hong Kong, Hong Kong SAR, and the Center for Transformative Garment Production, Hong Kong SAR.

\footnotesize Ning Guo and Fang Wan are with the Southern University of Science and Technology, Shenzhen, China, supported by the National Natural Science Foundation of China under Grant 62206119 and Shenzhen Long-Term Support for Higher Education at SUSTech under Grant 20231115141649002. Chaoyang Song is with the Mohamed bin Zayed University of Artificial Intelligence, Abu Dhabi, United Arab Emirates, supported by the National Natural Science Foundation of China under Grant 62473189.

\footnotesize Yixiang Jin, Yong A, Jun Shi, Dingzhe Li, Yong Yang are with the Samsung R\&D Institute China – Beijing (SRC-B), Beijing, China.

\footnotesize Kakeru Yamasaki, Takumi Kajiwara, Yuki Nakadera, Krati Saxena, and Tomohiro Shibata are with the Graduate School of Life Science and Systems Engineering, Kyushu Institute of Technology, Fukuoka, Japan

\footnotesize Chongkun Xia, Kai Mo, and Yanzhao Yu are with the Tsinghua University, Beijing, China.

\footnotesize Qihao Lin, and Binqiang Ma are with the Sun Yat-sen University, Guangzhou, China.

\footnotesize Uihun Sagong, JungHyun Choi, JeongHyun Park, Dongwoo Lee, Yeongmin Kim, and Myun Joong Hwang are with the Department of Mechanical and Information Engineering, University of Seoul, South Korea

\footnotesize Yusuke Kuribayashi, Naoki Hiratsuka, Solvi Arnold, and Kimitoshi Yamazaki are with Shinshu University, Matsumoto, Japan.

\footnotesize Daisuke Tanaka is with Tokyo Robotics Inc., Bunkyō, Japan.

\footnotesize Carlos Mateo-Agullo is with the Université de Bourgogne, Dijon, France.

\bibliographystyle{SageH}
\bibliography{references}

\end{document}